\documentclass[nohyperref]{article}

\usepackage{microtype}
\usepackage{graphicx}
\usepackage{subfigure}
\usepackage{booktabs} % for professional tables
\usepackage{float}
\usepackage{stfloats}
\usepackage{hyperref}

\usepackage[utf8]{inputenc}

\usepackage[arxiv]{arxiv}
\usepackage{amsmath, amsthm, booktabs}

\definecolor{blue}{HTML}{4d71a6}
\definecolor{green}{HTML}{2e7647}
\definecolor{brown}{HTML}{6d5959}
\definecolor{orange}{HTML}{DE9102}
\definecolor{red}{HTML}{ff4e33}

\definecolor{dmcyan}{HTML}{2BB2A6}
\definecolor{dmred}{HTML}{E1144B}
\definecolor{dmorange}{HTML}{E26A36}
\definecolor{dmraspberry}{HTML}{C72E57}
\definecolor{dmpurple}{HTML}{6932E6}
\definecolor{dmyellow}{HTML}{9CA352}

\hypersetup{
	colorlinks=true,       % false: boxed links; true: colored links
	linkcolor=dmcyan,        % color of internal links
	citecolor=dmcyan,        % color of links to bibliography
	filecolor=dmcyan,     % color of file links
	urlcolor=dmcyan         
}

\usepackage{amsmath,amsfonts,bm}

\def\1{\bm{1}}

\def\eps{{\epsilon}}

\DeclareMathAlphabet{\mathsfit}{\encodingdefault}{\sfdefault}{m}{sl}
\SetMathAlphabet{\mathsfit}{bold}{\encodingdefault}{\sfdefault}{bx}{n}

\def\gL{{\mathcal{L}}}

\def\gN{{\mathcal{N}}}
\def\gO{{\mathcal{O}}}

\def\gU{{\mathcal{U}}}

\def\sP{{\mathbb{P}}}

\def\sR{{\mathbb{R}}}
\def\sS{{\mathbb{S}}}

\def\sW{{\mathbb{W}}}
\def\sX{{\mathbb{X}}}

\newcommand{\E}{\mathbb{E}}

\DeclareMathOperator*{\argmin}{arg\,min}

\newcommand{\x}{\mathbf{x}}
\renewcommand{\d}{\mathrm{d}}

\let\bar\overline

\theoremstyle{remark}

\renewcommand{\x}{{\boldsymbol{x}}}
\renewcommand{\c}{{\boldsymbol{c}}}
\newcommand{\e}{{\boldsymbol{e}}}
\newcommand{\w}{{\boldsymbol{w}}}
\newcommand{\id}{\mathrm{d}}
\newcommand{\s}{s}
\newcommand{\tpi}{\tilde\pi}
\newcommand{\ppi}{\pi(\x)}
\newcommand{\energ}{E}

\newcommand{\mgn}{\textsc{MeshGNN}}
\everypar{\looseness=-1}

\begin{document}

\title{Diffusion Generative Inverse Design}
% \title{?}
\twocolumn[
\icmltitle{Diffusion Generative Inverse Design}

\icmlsetsymbol{dagger}{$\dagger$}

\begin{icmlauthorlist}
\icmlauthor{Marin Vlastelica}{mpi,dm}
\icmlauthor{Tatiana L\'{o}pez-Guevara}{dm}
\icmlauthor{Kelsey Allen}{dm}
\icmlauthor{Peter Battaglia}{dm}
\icmlauthor{Arnaud Doucet}{dm}
\icmlauthor{Kimberly Stachenfeld}{dm,cu}

\end{icmlauthorlist}

\icmlaffiliation{mpi}{Max Planck Institute for Intelligent Systems, T\"ubingen, Germany}
\icmlaffiliation{dm}{Google DeepMind, London, UK}
\icmlaffiliation{cu}{Columbia University, New York, NY}

\icmlcorrespondingauthor{Marin Vlastelica}{marin.vlastelica@tue.mpg.de} % Marin
\icmlcorrespondingauthor{Kimberly Stachenfeld}{stachenfeld@deepmind.com} % Kim
\icmlkeywords{Machine Learning, Diffusion, Generative Modelling, Learned Simulation, Graph Neural Networks, Fluid Dynamics}

\vskip 0.3in
]
\printAffiliationsAndNotice{} 

\begin{abstract}
Inverse design refers to the problem of optimizing the input of an objective function in order to enact a  target outcome.
For many real-world engineering problems, the objective function takes the form of a simulator that predicts how the system state will evolve over time, and the design challenge is to optimize the initial conditions that lead to a target outcome.
Recent developments in learned simulation have shown that graph neural networks (GNNs) can be used for accurate, efficient, differentiable estimation of simulator dynamics, and support high-quality design optimization with gradient- or sampling-based optimization procedures.
However, optimizing designs from scratch requires many expensive model queries, and these procedures exhibit basic failures on either non-convex or high-dimensional problems.
In this work, we show how denoising diffusion models (DDMs) can be used to solve inverse design problems efficiently and propose a particle sampling algorithm for further improving their efficiency. We perform experiments on a number of fluid dynamics design challenges, and find that our approach substantially reduces the number of calls to the simulator compared to standard techniques.
\end{abstract}

\section{Introduction}
Substantial improvements to our way of life hinge on devising solutions to engineering challenges, an area in which Machine Learning (ML) advances is poised to provide positive real-world impact. Many such problems can be formulated as designing an object that gives rise to some desirable physical dynamics (e.g. designing an aerodynamic car or a watertight vessel). Here we are using ML to accelerate this design process by learning both a forward model of the dynamics and a distribution over the design space.

Prior approaches to ML-accelerated design have used neural networks as a differentiable forward model for optimization \citep{challapalli2021inverse, christensen2020predictive, bombarelli2018Automatic}. 
We build on work in which the forward model takes the specific form of a GNN trained to simulate fluid dynamics \citep{allen2022phyical}.
Since the learned model is differentiable, design optimization can be accomplished with gradient-based approaches (although these struggle with zero or noisy gradients and local minima) or sampling-based approaches (although these fare poorly in high-dimensional design spaces).
Both often require multiple expensive calls to the forward model.
However, generative models can be used to propose plausible designs, thereby reducing the number of required calls \citep{forte2022inverse, zheng2020data, kumar2020inverse}.

\begin{figure*}[!t]
\centering
\includegraphics[width=0.96\textwidth]{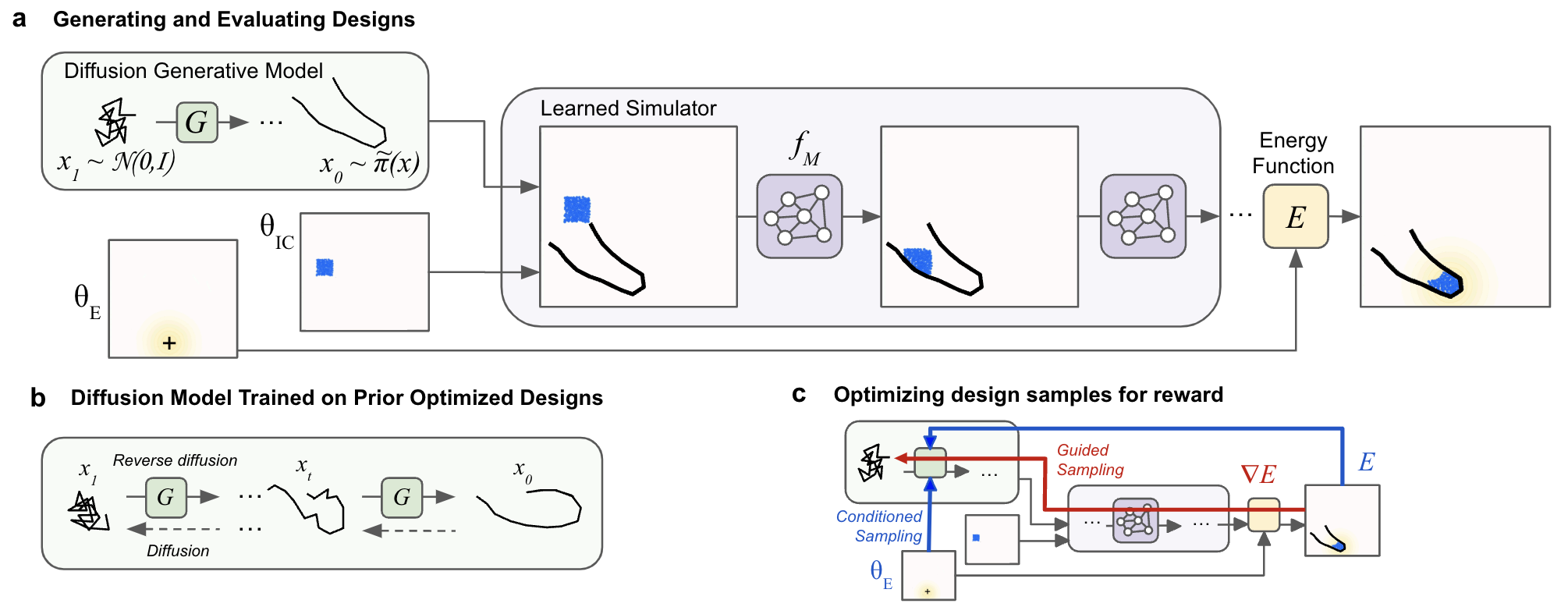}
\vspace{-0.3em}
\caption{(a) Given initial conditions governed by $\theta_\textrm{IC}$, energy function parameters $\theta_E$, and learned GNN dynamics model $f_M$, design samples $\x$ from the diffusion model are assigned a cost $E(\x)$. (b) Schematic of the DDM training (c) Gradients $\nabla E$ and conditioning set ($\theta_E$ and $E$) inform energy and conditional guidance, resp.}
\vspace{-0.5em}
\label{fig:schematic}
\end{figure*}

In this work, we use DDMs to optimize designs by sampling from a target distribution informed by a learned data-driven prior.
% application of diffusion generative models
DDMs have achieved extraordinary results in image generation  \cite{song2020denoising, song2020score, karras2022elucidating, ho2020denoising},
and has since been used to learn efficient planners in sequential decision making and reinforcement learning \citep{janner2022planning, ajay2022conditional}, sampling on manifolds \citep{bortoli2022riemannian} or constrained optimization formulations \citep{graikos2022diffusion}.
Our primary contribution is to consider DDMs in the setting of physical problem solving.
We find that such models combined with continuous sampling procedures enable to solve design problems orders of magnitude faster than off-the-shelf optimizers such as CEM and Adam.
This can be further improved by utilizing a particle sampling scheme to update the base distribution of the diffusion model which by cheap evaluations (few ODE steps) with a learned model leads to better designs in comparison to vanilla sampling procedures.
We validate our findings on multiple experiments in a particle fluid design environment.

\section{Method}

Given some task specification $\c$, we have a target distribution of designs $\pi(\x)$ which we want to optimize w.r.t. $\x$. To simplify notation, we do not emphasize the dependence of $\pi$ on $\c$.
This distribution is a difficult object to handle, since a highly non-convex cost landscape might hinder efficient optimization.
We can capture prior knowledge over `sensible' designs in form of a prior distribution $p(\x)$ learned from existing data.
Given a prior, we may sample from the distribution
\begin{equation}\label{eq:truetarget}
\tilde\pi(\x) \propto  p(\x) \pi(\x ),
\end{equation}
which in this work is achieved by using a diffusion method with guided sampling.
The designs will subsequently be evaluated by a learned forward model comprised of a pretrained GNN simulator and a reward function \cite{allen2022phyical,pfaff2021learning,sanchez2020learning} (see \autoref{app:gnn}).

Let  $E: \sX \mapsto \sR$  be the cost (or ``energy'') of a design $\x \in \sX$ for a specific task $\c$ under the learned simulator (dependence of $E$ on $\c$ is omitted for simplicity). The target distribution of designs $\pi(\x)$ is defined by the Boltzmann distribution
\begin{equation}
   \pi(\x) := \frac{1}{Z} \exp\left (-\frac{\energ(\x)}{\tau} \right ),
\end{equation}
where $Z$ denotes the unknown normalizing constant and $\tau$ a temperature parameter. As $\tau \rightarrow 0$, this distribution concentrates on its modes, that is on the set of the optimal designs for the cost $E^\c(\x)$. Direct methods to sample from $\ppi$ rely on expensive Markov chain Monte Carlo techniques or variational methods minimizing a reverse KL criterion.

We will rely on a data-driven prior learned by the diffusion model from previous optimization attempts.
We collect optimization trajectories of designs for different task parametrizations $\c$ using Adam~\citep{adam} or CEM~\citep{rubinstein1999cross} to optimize $\x$.
Multiple entire optimization trajectories of designs are included in the training set for the generative model, providing a mix of design quality.
These optimization trajectories are initialized to flat tool(s) below the fluid (see \autoref{app:fig:environments}), which can be more easily shaped into successful tools than a randomly initialized one. Later, when we compare the performance of the DDM to Adam and CEM, we will be using randomly initialized tools for Adam and CEM, which is substantially more challenging.

\subsection{Diffusion generative models}
We use DDMs to fit $p(\x)$ \cite{ho2020denoising,song2020score}. The core idea is to initialize using training data $\x_0 \sim p$, captured by a diffusion process $(\x_t)_{t\in [0,1]}$ defined by 
\begin{equation}\label{eq:forwardSDE}
    \id \x_t = -\beta_t \x_t \id t +\sqrt{2\beta_t}\id \w_t,
\end{equation}
where $(\w_t)_{t\in[0,1]}$ denotes the Wiener process. We denote by $p_t(\x)$ the distribution of $\x_t$ under (\ref{eq:forwardSDE}). For $\beta_t$ large enough, $p_1(\x) \approx \mathcal{N}(\x;0,I)$. The time-reversal of (\ref{eq:forwardSDE}) satisfies
\begin{equation}\label{eq:reverseSDE}
 %   \id \x_t = [f(\x_t, t) - g(t)^2 \nabla_{\x} \log p_t(\x_t)]\id t + g(t)\id \w^{-}_t,
    \id \x_t = -\beta_t[\x_t + 2\nabla_{\x} \log p_t(\x_t)]\id t +\sqrt{2\beta_t} \id \w^{-}_t,
\end{equation}
where $(\w^{-}_t)_{t\in [0,1]}$  is a Wiener process when time flows backwards from $t=1$ to $t=0$, and $\id t$ is an infinitesimal negative timestep. By initializing (\ref{eq:reverseSDE}) using $\x_1 \sim p_1$, we obtain $\x_0 \sim p$. In practice, the generative model is obtained by sampling an approximation of (\ref{eq:reverseSDE}), replacing $p_1(\x)$ by $\mathcal{N}(\x;0,I)$ and the intractable score $\nabla_x \log p_t(\x)$ by $s_\theta(\x,t)$. 
The score estimate $s_\theta(\x,t)$ is learned by denoising score matching, i.e. we use the fact that $\nabla_x \log p_t(\x)=\int \nabla_x \log p(\x_t|\x_0) p(\x_0|\x_t) \id \x_0$ where $p(\x_t|\x_0)=\mathcal{N}(\x_t;\sqrt{\alpha_t}\x_0,\sqrt{1-\alpha_t}I)$ is the transition density of (\ref{eq:forwardSDE}), $\alpha_t$ being a function of $(\beta_s)_{s\in[0,t]}$ \citep{song2020score}. It follows straightforwardly that the score satisfies $\nabla_x \log p_t(\x)=-\mathbb{E}[\epsilon|\x_t=\x]/\sqrt{1-\alpha_t}$ for  $\x_t = \sqrt{\alpha_t} \x_0  + \sqrt{1-\alpha_t}\eps$. We then learn the score by minimizing 
\begin{align}
\gL(\theta) = \E_{\x_0 \sim p, t \sim \gU(0,1), \eps \sim \gN(0,I)}\|\eps_\theta(\x_t, t) - \eps  \|^2,
\end{align}
where $\eps_\theta(\x,t)$ is a denoiser estimating $\mathbb{E}[\epsilon|\x_t=\x]$. 
The score function $\s_\theta(\x, t) \approx \nabla_\x \log p_t(\x)$ is obtained using
\begin{align}\label{eq:eps-score-rel}
    \s_\theta(\x, t) = - \frac{\eps_\theta(\x,t)}{\sqrt{1-\alpha_t}}.
\end{align}
Going forward, $\nabla$ refers to $\nabla_\x$ unless otherwise stated.
We can also sample from $p(\x)$ using an ordinary differential equation (ODE) developed in \citep{song2020score}. 
%This will adopted here.

Let us define  $\bar \x_t = \x_t /\sqrt{\alpha_t}$  and $\sigma_t = \sqrt{1-\alpha_t}/\sqrt{\alpha_t}$. Then by initializing $\x_1 \sim \mathcal{N}(0,I)$, equivalently $\bar{\x}_1 \sim \mathcal{N}(0,\alpha_1^{-1}I)$ and solving backward in time 
\begin{equation}\label{eq:ODE}
    \d \bar \x_t= \eps_\theta^{(t)} \left ( \frac{\bar \x_t}{\sqrt{\sigma^2_t + 1}} \right ) \d\sigma_t,
\end{equation}
then $\x_0=\sqrt{\alpha_t}~\bar \x_0$ is an approximate sample from $p(\x)$.

\subsection{Approximately sampling from target $\tilde\pi(\x)$} \label{sec:posterior-sampling}

We want to sample $\tilde\pi(\x)$ defined in (\ref{eq:truetarget}) where $p(\x)$ can be sampled from using the diffusion model.
We describe two possible sampling procedures with different advantages for downstream optimization.

\textbf{Energy guidance.} Observe that
\begin{equation*}
    \tpi_t(\x_t)=\int \tpi(\x_0)p(\x_t|\x_0)\id \x_0,
\end{equation*}
and the gradient satisfies
\begin{equation*}
      \nabla \log \tpi_t(\x_t) = \nabla \log p_t(\x_t) +\nabla \log \pi_t(\x_t), 
\end{equation*}
where $\pi_t(\x_t)=\int \pi(\x_0) p(\x_0|\x_t) \id \x_0$. We approximate this term by making the approximation
\begin{equation}\label{eq:pit-approx}
\begin{array}{rcl}
     \hat \x(\x_t, t)  &=&  \underbrace{\left(\frac{\x_t - \sqrt{1 - \alpha_t} \epsilon_\theta(\x_t, t)}{\sqrt{\alpha_t}}\right)}_{\text{`` estimated } \x_0 \text{''}},\\
    \pi_t(\x_t) &\approx& \pi(\hat\x_t(\x_t, t)).
\end{array}
\end{equation}
Now, by (\ref{eq:eps-score-rel}), and the identity $\nabla \log \pi(\x) =  -\tau^{-1} \nabla \energ(\x)$, we may change the reverse sampling procedure by a modified denoising vector
\begin{align}\label{eq:energy-guidance}
    \tilde \eps_\theta(\x_t, t) = \eps_\theta(\x_t, t) + \lambda \tau^{-1} \sqrt{1-\alpha_t} \nabla \energ (\hat\x(\x_t, t)),
\end{align}
with $\lambda$ being an hyperparameter. We defer the results on energy guidance \autoref{app:sec:energy-guidance}.

\textbf{Conditional guidance.} Similarly to \emph{classifier-free} guidance \cite{ho2022classifier}, we explore conditioning on cost (energy) $\e$ and task $\c$.
A modified denoising vector in the reverse process follows as a combination between the denoising vector of a conditional denoiser $\eps_\phi$ and unconditional denoiser $\eps_\theta$
\begin{align}
    \tilde \eps(\x_t, \c, \e, t) = (1+\lambda)\eps_\phi(\x_t, \c, \e, t) - \lambda \eps_\theta(\x_t, t),
\end{align}
where $\eps_\phi$ is learned by conditioning on $\c$ and cost $\e$ from optimization trajectories.
In our experiments we shall choose $\c$ to contain the design cost percentile and target goal destination $\theta_E$ for fluid particles (\autoref{fig:schematic}c).

\subsection{A modified base distribution through particle sampling}\label{sec:particle-search}

Our generating process initializes samples at time $t=1$ from $\gN(\x; 0, I)\approx p_1(\x)$. The reverse process with modifications from \autoref{sec:posterior-sampling} provides approximate samples from $\tpi(\x)$ at $t=0$. However, as we are approximately solving the ODE of an approximate denoising model with an approximate cost function, this affects the quality of samples with respect to $\energ$\footnote{Ideally at test time we would evaluate the samples with the ground-truth dynamics model, but we have used the approximate GNN model due to time constraints on the project.}.
Moreover, ``bad'' samples from $\gN(\x; 0, I)$ are hard to correct by guided sampling.

\scalebox{0.75}{}
\begin{algorithm}[tb]
\begin{algorithmic}[2]
        {\scriptsize {
        \STATE \textbf{input} energy function $\energ$, diffusion generative model $p_\theta$, temperature $\tau$, noise scale $\sigma$, rounds $K$.
		\STATE $\sS^0_1 = \{\x_1^i \}_{i=1}^N$ for $\x_1^i \overset{\textup{i.i.d.}}{\sim} \gN(0, I) $ % \COMMENT{Initial particles.}
		\STATE $\sS_0 = \emptyset$, $\sS_1 = \emptyset$ \#  $t=0$  and $t=1$ sample sets
		\FOR{ $k \in \{0 \dots K\}$}
		\STATE Compute $\sS_0^k = \{\x_0^i \}_{i=1}^N$ from $\sS_1^k$ by solving reverse ODE in Eq. (\ref{eq:ODE}).
		\STATE $\sS_0 =  \sS_0 \cup \sS_0^k$, $\sS_1 = \sS_1 \cup \sS_1^k$

		\STATE Compute normalized importance weights \\ $\sW = \Big \{ w~|~  w \propto \exp\big(-\frac{\energ(\x_0)}{\tau} \big),~ \x_0 \in \sS_0  \Big \}$
      	\STATE Set $\bar\sS_1^{k+1} =\{\bar{\x}_1^i\}_{i=1}^{|\sS_1|}$ for $\bar{\x}_1^i  \overset{\textup{i.i.d.}}{\sim}  \sum_{i=1}^{|\sS_1|} w^i \delta_{\x^i_1}(\x)$
     	\STATE Set $\sS^{k+1}_1 =\{\tilde{\x}_1^i\}_{i=1}^{|\sS_1|}$ for $\tilde{\x}_1^i \sim \gN(\x; \bar{\x}_1^i, \sigma^2 I)$ % \# small perturbation

		% \STATE Sample \\ $\sS^{k+1}_1 = \left \{ \x ~ | ~ \x \sim \gN(\x; \x^i_1, \sigma^2 I), ~ \x^i_1 \in \bar\sS^{k+1}_1 \right \}$ % \# small perturbation
		\ENDFOR
		\STATE \textbf{return} $\argmin_{\x \in \sS_0} \energ(\x)$}}
	\end{algorithmic}
\caption{Particle optimization of base distribution.}
\label{algo:particle-search}
\end{algorithm}
\vspace{-2em}

To mitigate this, instead of using samples from $\gN(\x; 0, I)$ to start the reverse process of $\tilde\pi(\x)$, we use a multi-step particle sampling scheme which evaluates the samples $\{\x^i_1\}_{i=1}^N$ by a rough estimate of the corresponding $\{\x^i_0\}_{i=1}^N$ derived from a few-step reverse process and evaluation with $\energ$.
The particle procedure relies on re-sampling from a weighted particle approximation of $\pi(\x)$ and then perturbing the resampled particles, see \ref{algo:particle-search}. This heuristic does not provide samples from $\tilde{\pi}$ but we found that it provides samples of lower energy samples across tasks. However, with $N$ samples in each of the $k$ rounds, it still requires $\gO(Nk)$ evaluations of $\energ$, which may be prohibitively expensive depending on the choice of $\energ$.

\section{Experiments} \label{sec-experiments}

We evaluate our approach on a 2D fluid particle environment with multiple tasks of varying complexity, which all involve shaping the tool (\autoref{fig:schematic}a, black lines) such that the fluid particles end up in a region of the scene that minimizes the cost $\energ$ \cite{allen2022phyical}. 
As baselines, we use the Adam optimizer combined with a learned model of the particles and the CEM optimizer which also optimizes with a learned model.
For all experiments we have used a simple MLP as the denoising model and GNN particle simulation model for evaluation of the samples.

\begin{figure}[t]
    \centering
    \includegraphics[width=0.5\textwidth]{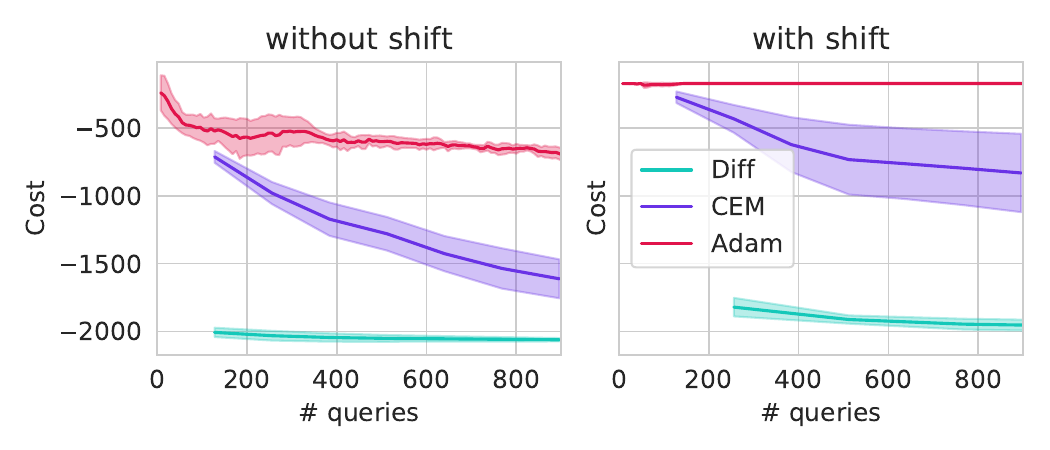}
    \vspace{-7mm}
    \caption{Performance of the different optimization methods in the angle optimization task. We observe that the diffusion generative model requires a small number of model queries, whereas Adam in comparison requires many expensive model queries.}
    \label{fig:simple-task}
\end{figure}

\begin{figure}[b]
    \centering
    \includegraphics[width=0.4\textwidth]{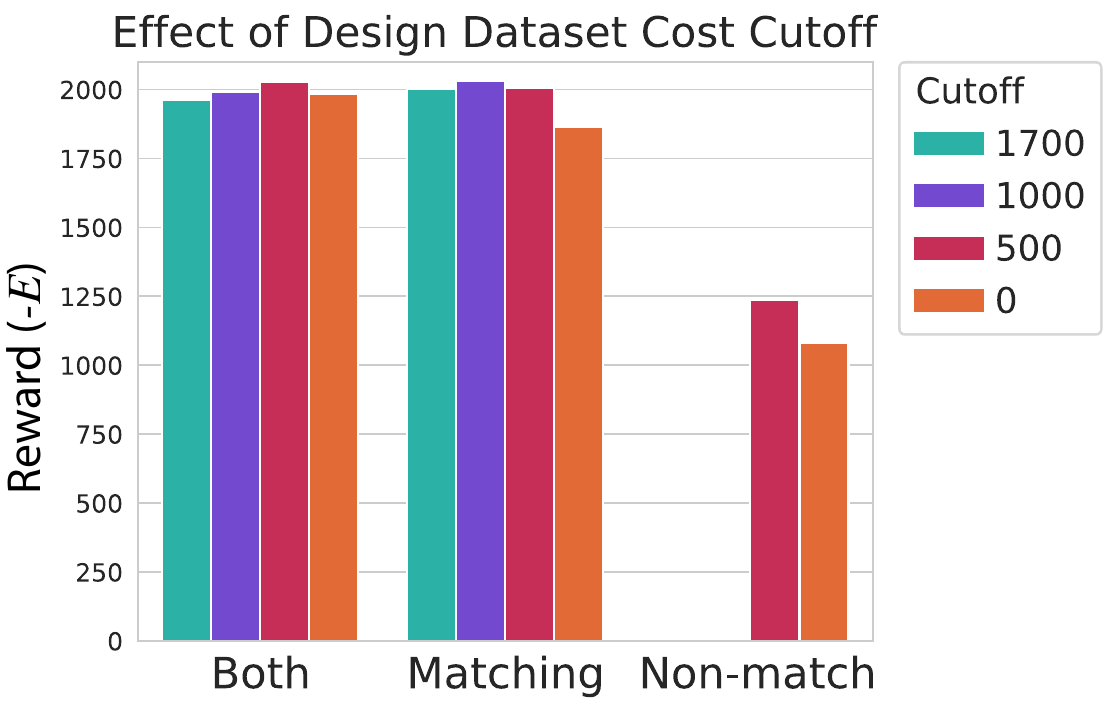} 
    \caption{ 
    Generative models were trained on a dataset of designs produced from CEM- and Adam-optimized designs on either of two tasks Matching, Non-matching, or Both. Designs in the train dataset were filtered to have costs below the specified cutoff. }
   \label{fig:data-quality}
\end{figure}

The first task is a simple ``Contain'' task with a single source of water particles,  a goal position above the floor specified by $\c = (x,y)$, and an articulated tool with 16 joints whose angles are optimized to contain the fluid near the goal (see~\citet{allen2022phyical}).
In \autoref{fig:simple-task}a, we see that both the Adam optimizer and CEM are able to optimize the task. However with training a prior distribution on optimization trajectories and guided sampling we are able to see the benefits of having distilled previous examples of optimized designs into our prior, and achieve superior performance with fewer samples in unseen tasks sampled in-distribution.

If we modify the task by introducing a parameter controlling $x,y$ shift parameter, we observe that Adam fails. This is because there are many values of the shift for which the tool makes no contact with the fluid (see \autoref{fig:simple-task}b), and therefore gets no gradient information from $\energ$.
We provide results for more complex tasks in \autoref{app:sec:particle-environments} (\autoref{app:fig:environments}). Overall, we find that this approach is capable of tackling a number of different types of design challenges, finding effective solutions when obstacles are present, for multi-modal reward functions, and when multiple tools must be coordinated.

\subsection{Dataset quality impact}

We analyzed how the model performs with conditional guidance when trained on optimization trajectories of CEM that optimize the same task (matching), a different task (non-matching), or a mix of tasks (\autoref{fig:data-quality}).
The two tasks were ``Contain'' (described above) and ``Ramp'' (transport fluid to the far lower corner).
Unsurprisingly, the gap between designs in the training dataset and the solution for the current task has a substantial impact on performance.
We also control the quality of design samples by filtering out samples above a certain cost level for the $\c$ with which they were generated.
Discarding bad samples from the dataset does not improve performance beyond a certain point: best performance is obtained with some bad-performing designs in the dataset. Intuitively, we believe this is because limiting the dataset to a small set of optimized designs gives poor coverage of the design space, and therefore generalizes poorly even to in-distribution test problems. 
Further, training the generative model on samples from optimization trajectories of only a non-matching task has a substantial negative impact on performance.
We expect the energy guidance not to suffer from the same transfer issues as conditional guidance, since more information about the task is given through the energy function.
Since we indeed obtain data to fit $p(\x)$ from Adam and CEM runs, why is diffusion more efficient?
We discuss this in \autoref{app:sec:results-discussion}. 

\begin{figure}[b]
    \centering
    \includegraphics[width=0.8\linewidth]{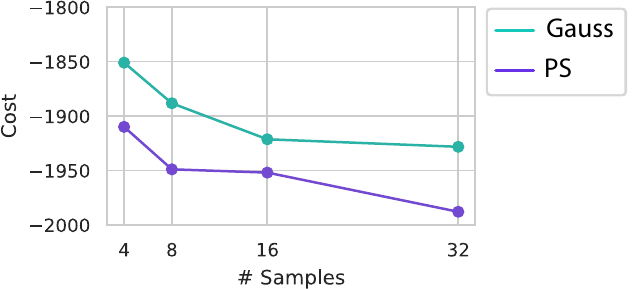}    
    \caption{Gaussian base distribution vs. particle sampling (PS).}
    \label{fig:particle-sampling}
\end{figure}

\subsection{Particle optimization in base distribution}

We also observe performance improvements by using the particle search scheme from \autoref{sec:particle-search}, see \autoref{fig:particle-sampling}.
We hypothesize that the reason for this is because of function approximation. 
Since we are dealing with approximate scores, it is hard for the learned generative model to pinpoint the optima, therefore a sampling-based search approach helps.
We note that the evaluation of a sample with $\energ$ requires solving the ODE for sample $\x_1$, we use a 1-step reverse process making use of the relation in (\ref{eq:pit-approx}).
Consequently, we can expect that linearizing the sampling will improve the particle search.

\section{Conclusion}

In this work we have demonstrated the benefits of using diffusion generative models in simple inverse designs tasks where we want to sample from high probability regions of a target distribution $\pi(\x)$ defined via $\energ$, while having access to optimization data. 
We analyzed energy-based and conditional guidance where the energy function involves rolling out a GNN. We find that energy guidance is a viable option, but conditional guidance works better in practice, and that performance depends heavily on the generative model's training data.
Finally, we have introduced particle search in the base distribution as a means to improve quality of the samples and demonstrated this on multiple tasks.

\section{Acknowledgments}
We would like to thank Conor Durkan, Charlie Nash, George Papamakarios, Yulia Rubanova, and Alvaro Sanchez-Gonzalez for helpful conversations about the project. We would also like to thank Alvaro Sanchez-Gonzalez for comments on the manuscript. 

\everypar{\looseness=-2}

\bibliography{main}
\bibliographystyle{icml2022}

\newpage

\appendix

\twocolumn[
\centering
{\Large \textbf{Appendix for Diffusion Generative Inverse Design}}
\vspace{2em}

]

\begin{figure*}[b]
    \centering
    \includegraphics[width=.7\linewidth]{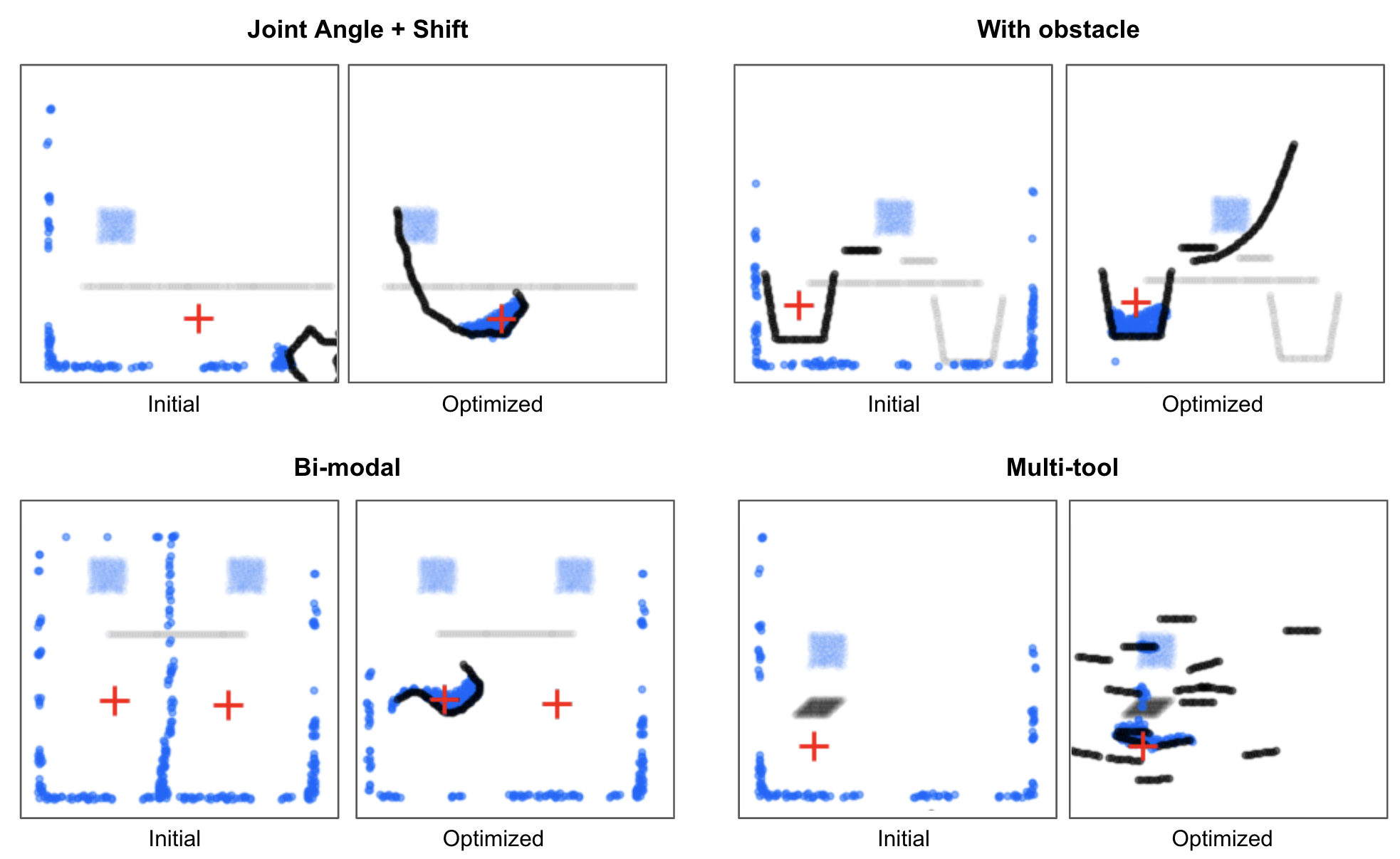}
    \caption{Examples of guided-diffusion generated designs for different tasks $\c$ that we consider in this work.
    The initial designs start off completely at random, and the optimized ones solve the task.}
    \label{app:fig:environments}
\end{figure*}

\begin{figure*}[t]
    \centering
    \includegraphics[width=.5\linewidth]{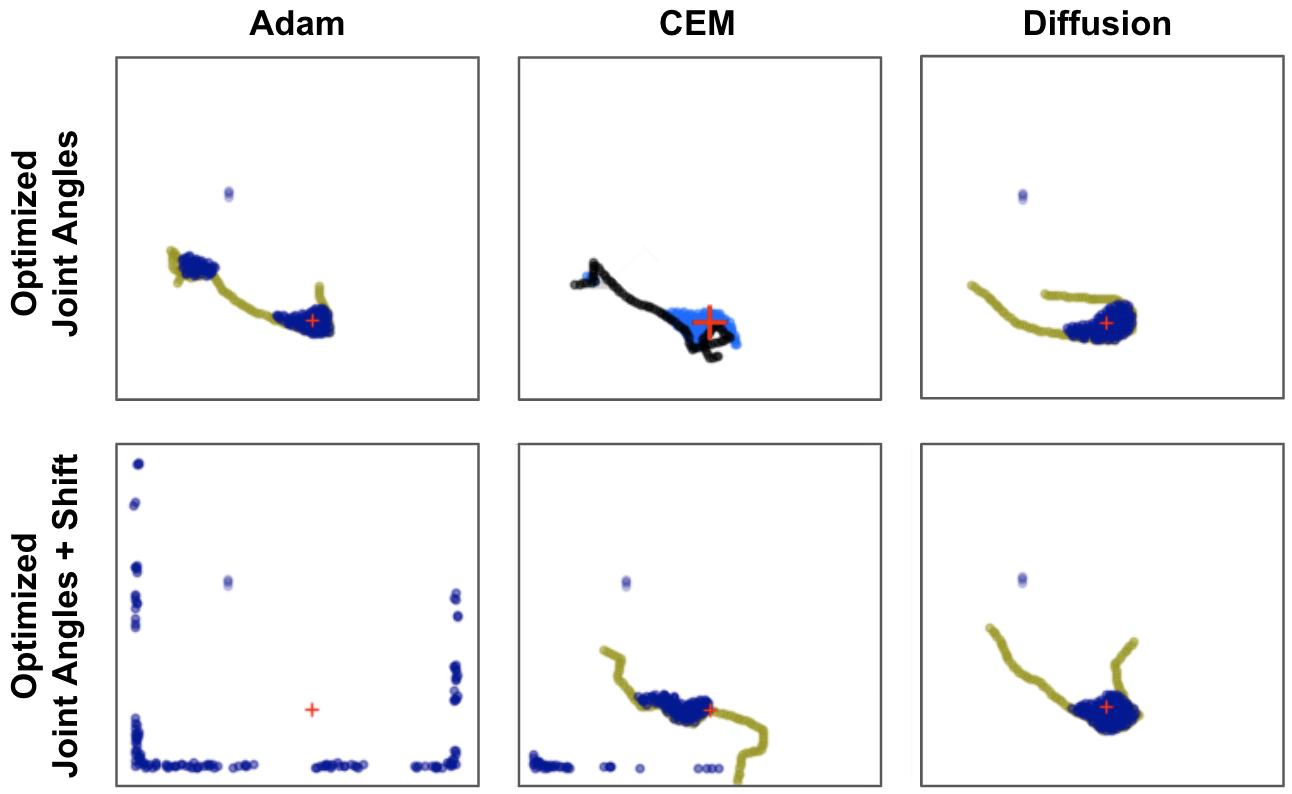}
    \caption{Examples of designs optimized with Adam, CEM, and guided diffusion using the generative model. 
    Designs are initialized as random joint angles. Each design shown is the top scoring design for that optimizer, evaluated under the learned model, after having been trained on 1000 calls to the simulator (the limit of the $x$-axis in \autoref{fig:simple-task}).}
    \label{app:fig:adam-cem-diff}
\end{figure*}

\section{Learned simulation with graph neural networks}\label{app:gnn}
As in \citet{allen2022phyical}, we rely on the recently developed \mgn{} model \citep{pfaff2021learning}, which is an extension of the GNS model for particle simulation \citep{sanchez2020learning}.
\mgn{} is a type of message-passing graph neural network (GNN) that performs both edge and node updates \citep{battaglia2018relational,gilmer2017neural}, and which was designed specifically for physics simulation.

We consider simulations over physical states represented as graphs $G\in\mathcal{G}$.
The state $G = (V, E)$ has nodes $V$ connected by edges $E$, where each node $v\in V$ is associated with a position $\mathbf{u}_v$ and additional dynamical quantities $\mathbf{q}_v$. 
In this environment, each node corresponds to a particle and edges are computed dynamically based on proximity.

The simulation dynamics on which the model is trained are given by a ``ground-truth'' simulator which maps the state $\mathcal{G}^t$ at time $t$ to the state $\mathcal{G}^{t+1}$ at time $t+\Delta t$.
The simulator can be applied iteratively over $K$ time steps to yield a trajectory of states, or a ``rollout,'' which we denote $(G^{t_0}, ..., G^{t_K})$.
The GNN model $M$ is trained on these trajectories to imitate the ground-truth simulator $f_S$.
The learned simulator $f_M$ can be similarly applied to produce rollouts $(\tilde{G}^{t_0}, \tilde{G}^{t_1},  ..., \tilde{G}^{t_K})$, where $\tilde{G}^{t_0} = G^{t_0}$ represents initial conditions given as input.

\section{Further discussion on results}\label{app:sec:results-discussion}

We trained the diffusion model on data generated from Adam and CEM optimization runs and noticed an improvement over Adam and CEM on the evaluation tasks.
The reason for this is that both Adam and CEM need good initializations to solve the tasks efficiently, for example in \autoref{fig:simple-task} for each run the initial design for Adam has uniformly sampled angles and $x,y$ coordinates within the bounds of the environment, which would explain why we obtain worse results \emph{on average} for Adam than \citet{allen2022phyical}. 
Similarly, for CEM we use a Gaussian sampling distribution which is initialized with zero mean and identity covariance.
If most of the density of the initial CEM distribution is not concentrated near the optimal design, then CEM will require many samples to find it.

In comparison, the diffusion model learns good initializations of the designs through $p(\x)$ which can further be improved via guided sampling, as desired.

\section{Particle environments}\label{app:sec:particle-environments}

For evaluation, we consider similar fluid particle-simulation environments as in \citet{allen2022phyical}.
The goal being to design a `tool' that brings the particles in a specific configuration. 
We defined the energy as the radial basis function 
\begin{equation*}
    E(\x) = \sum_{p\in\sP} \exp\left (- \frac{\|\x^t_p - \x_p^\star\|}{\sigma} \right),
\end{equation*}
where $x^t_p$ are the coordinates of particle $p$ after rolling out the simulation with the model $f_M$ with initial conditons $\theta_\mathrm{IC}$ and parameters $\theta_E$.
Note that the energy is evaluated on the last state of the simulation, hence $\nabla E$ needs to propagate through the whole simulation rollout.

In additon to the ``Contain'' environments described in \autoref{sec-experiments}, we provide further example environments that we used for evaluation with the resulting designs from guided diffusion can be seen in \autoref{app:fig:environments}.

The task with the obstacle required that the design is fairly precise in order to bring the fluid into the cup, this is to highlight that the samples found from the diffusion model with conditional guidance + particle sampling in base distribution are able to precisely pinpoint these types of designs.

For the bi-modal task, where we have two possible minima of the cost function, we are able to capture both of the modes with the diffusion model.

In the case where we increase the dimensionality of the designs where we have $x,y$ shift parameters for each of the tool joints, and the tools are disjoint, the diffusion model is able to come up with a parameterization that brings the particles in a desired configuration.
However, the resulting designs are not robust and smooth, indicating that further modifications need to be made in form of constraints or regularization while guiding the reverse process to sample from  $\tilde\pi(\x)$.

\begin{figure}
    \centering
    \includegraphics[width=0.4\textwidth]{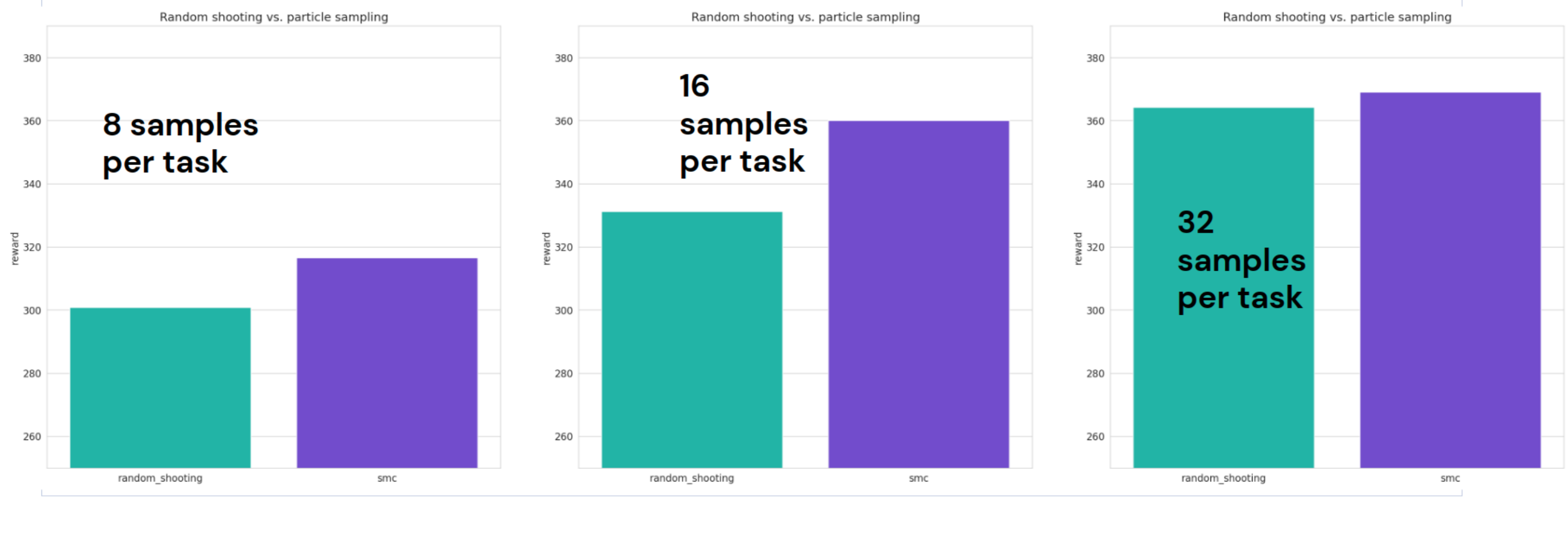}    
    
    {\color{dmcyan}\rule[.5ex]{1em}{.2em}} Gauss \quad {\color{dmpurple}\rule[.5ex]{1em}{.2em}} PS    
    \caption{Sampling (random search) from $p_1(\x)$ and particle sampling in the bi-modal environment. We observe that even after increasing the number of samples,  particle search further improves performance with same number of samples.}
\label{app:fig:multiple-blobs}
\end{figure}

\section{Discussion on choice of guidance}

As we will see in \autoref{sec-experiments}, conditional guidance with corrected base distribution sampling tends to work better than energy guidance.
In cases where the gradient of the energy function is expensive to evaluate, an energy-free alternative might be better, however this requires learning a conditional model, i.e. necessitates access to conditional samples.

\section{Energy guidance}\label{app:sec:energy-guidance}

\begin{figure}
    \centering
    \includegraphics[width=0.4\textwidth]{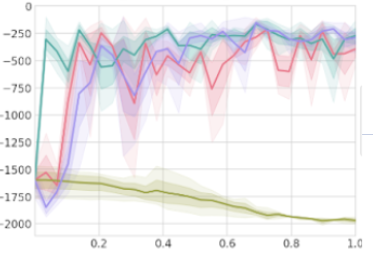}    
    
    {\color{dmcyan}\rule[.5ex]{1em}{.2em}}  \emph{linear}  \quad 
    {\color{dmred}\rule[.5ex]{1em}{.2em}} \emph{linear-unit} \quad {\color{dmyellow}\rule[.5ex]{1em}{.2em}} \emph{linear-norm} \quad
    {\color{dmpurple}\rule[.5ex]{1em}{.2em}} \emph{cvx}
    \caption{Performance of energy guidance depending on guidance scale $\lambda$ (x axis) for different modifications to score of noised marginal.}
    \label{app:fig:energy_performance}
\end{figure}

We have found that using the gradient of the energy function as specified in equation~(\ref{eq:energy-guidance}) is a viable way of guiding the samples, albeit coming with the caveat of many expensive evaluations, see \autoref{app:fig:energy_performance}.
The guidance is very sensitive to the choice of the scaling factor $\lambda$, in our experiments we have found that a smaller scaling factor with many integration steps achieves better sample quality.
Intuitively, this follows from the fact that it is difficult to guide `bad' samples in the base distribution $p_1(\x)$, which motivates the particle energy-based sampling scheme introduced in algorithm \ref{algo:particle-search}.

Further, we have looked at how to combine the gradient of the energy with the noised marginal score. We have found that re-scaling to have the same norm as the noised marginal improves the stability of guidance, as shown in $\autoref{app:fig:energy_performance}$.
Here, we have analyzed multiple functions with which we combined the energy gradient and the noised marginal score, we have looked at the following variants:
\begin{itemize} 
\item \emph{linear} - simple linear addition of $\lambda \nabla \energ$.

\item \emph{linear-unit} - linear addition of $\lambda \frac{\nabla \energ}{\|\nabla \energ\|}$.

\item \emph{cvx} - convex combination of $\nabla \energ$ and $\epsilon_\theta$.

\item \emph{linear-norm} -   linear addition of $\lambda \frac{\nabla \energ \|\epsilon_\theta\|}{\|\nabla \energ\|}$
\end{itemize}

\onecolumn

\begin{table}[t]
\begin{center}
\begin{tabular}{ r | c c c c c}
\toprule
	            &	Contain	 &	Ramp  &	With obstacle & Bi-modal  & Multi-tool	\\
\midrule
Environment size	&	1x1	     &	1x1   &	1x1           &	1x1       &	1x1	\\
Rollout length	    &	150	     &	150	  &	150           & 150	      &	150	\\
\\
Initial fluid box(es) & & & & & \\
\textit{left}	&   0.2   & 0.2   & 0.45  & 0.25, 0.65    & 0.2 \\
\textit{right}	&   0.3   & 0.3   & 0.55  & 0.35, 0.75    & 0.3 \\
\textit{bottom} &	0.5   &	0.5   & 0.5   &	0.5           & 0.5 \\
\textit{top}	&	0.6   &	0.6   & 0.6   &	0.6           & 0.6 \\
\\
Reward sampling box & & & & & \\
\textit{left}	    &	 0.4   & 0.8   &  0.2  & 0.25, 0.65 & 0.2 \\
\textit{right}	    &	 0.6   & 1.0   &  0.2  & 0.35, 0.75 & 0.3 \\
\textit{bottom}	    &	 0.1   & 0.0   &  0.2  & 0.1        & 0.2 \\
\textit{top}	    &	 0.3   & 0.2   &  0.2  & 0.2        & 0.5 \\
\\
Reward $\sigma$	    &	 0.1   & 0.1   & 0.05   & 0.1   & 0.1   \\
\\
\# tools	        &	1           	 &	1	            &	1   		  & 1             & 16 \\
\# joint angles	    &	16	             &	16	            &	16            & 16            & 1 \\
Design parameters   &	joint angles,     &	joint angles	&	joint angles, &	joint angles, & shift\\
            	&	shift (optional) &	             	&	shift		  & shift         & \\
\\
Tool position (left) &   [0.15, 0.35] &	[0.15, 0.35] &	[0.25, 0.35] & [0.3, 0.6] & [0.15--0.2, 0.35--0.4] \\
\\
Tool Length	        &	0.8	&	0.8	&	0.6 & 0.4 & 0.1\\
\\
Additional obstacles & --- & --- & barrier halfway & --- & ---\\
& & & between cup and fluid, & & \\
& & & cup around goal & & \\
% TODO: find parameters
\bottomrule
\end{tabular}
\end{center}
\caption{Task Parameters.}
\label{tab:2dfluids_tasks}
\end{table}

\end{document}